\documentclass{article}
\usepackage[utf8]{inputenc}
\usepackage{xcolor}
\usepackage{float}
\usepackage[justification=centering]{caption}
\usepackage[margin=0.5in]{geometry}
\usepackage[pdftex]{graphicx}
\graphicspath{{\Documents}}     
\DeclareGraphicsExtensions{.pdf,.jpeg,.png}
\usepackage{subfigure}
\usepackage{titlesec}
\usepackage{amssymb,amsmath}
\usepackage{moreverb}
\usepackage{multirow}
\usepackage[utf8]{inputenc}
\usepackage{natbib}
\usepackage{multicol}

\title{{Non-linear Functional Modeling using Neural Networks}}

\begin{multicols}{2}
\author{Aniruddha Rajendra Rao\\
Department of Statistics\\
Pennsylvania State University, USA\\
\vspace{1cm}
arr30@psu.edu\\
\and
Matthew Reimherr*\\
Department of Statistics\\
Pennsylvania State University, USA\\
mreimherr@psu.edu}
\end{multicols}

\date{}

\begin{document}

\maketitle
\vspace{-1.5cm}
\hspace{0.35cm}{Keywords:}   Functional Data Analysis, Deep Learning, Functional Regression, Functional Neural Network.

\abstract{We introduce a new class of non-linear models for functional data based on neural networks. Deep learning has been very successful in non-linear modeling, but there has been little work done in the functional data setting. We propose two variations of our framework: a functional neural network with continuous hidden layers, called the Functional Direct Neural Network (FDNN), and a second version that utilizes basis expansions and continuous hidden layers, called the Functional Basis Neural Network (FBNN). Both are designed explicitly to exploit the structure inherent in functional data. To fit these models we derive a functional gradient based optimization algorithm. The effectiveness of the proposed methods in handling complex functional models is demonstrated by comprehensive simulation studies and real data examples.}

\section{Introduction}\label{sec1}

In today's world, repeated measurements taken over time or space are collected across a variety of different areas of industry, government, and academia. For instance, health monitoring of patients, daily weather information, and stock prices, to name only a few. In such cases, the data of interest changes continuously over time.  
One approach to modeling such data is functional data analysis \citep{ramsay1997functional,FDA}.

Functional Data Analysis (FDA) is a branch of statistics that enables researchers to analyze and model relationships between variables measured over a particular domain, such as time or space \citep{ramsay1997functional,FDA,book1,10,hor}. FDA has been a rich area of research in the last few decades. The advantage FDA has over multivariate data analysis is that FDA enables smoothing and interpolation procedures that can yield functional representations of a finite set of observations. Also, FDA has applicability to data that are recorded irregularly over time or some other domain.


Functional regression can mainly be classified into three major categories according to the role played by the functional data in each model: scalar response and functional predictors (scalar-on-function regression); functional response and scalar predictors (function-on-scalar regression); and functional response and functional predictors (function-on-function regression). This paper concentrates on the first case, that is the scalar-on-function regression. Even though Functional Regression, especially scalar-on-function regression, is perhaps one of the most thoroughly researched topics within the broader literature on FDA, there is still a great need for superior non-linear models. It is not only a very active area for research, but also is very challenging with functional data due to the {\it curse of dimensionality} \citep{cd2010}. 


Some common ways of dealing with non-linearity in functional data are by using the generalized additive model, also known as Continuously Additive Model (CAM) \citep{McLean,articlel3,articlel4,articlel1,articlel2,reimherr2017optimal}, Quadratic Model \citep{quad}, Single Index Model \citep{s1,EILERS2009196,bookr1}, and Multiple Index Model \citep{articlem1,articlem2,articlem3}. The single index model has the outcome related to a linear functional of the predictors through a non-linear transformation, whereas the multiple index model is analogous but with multiple linear projections. In these models, you can either choose to capture the complex non-linear relations or the higher order interactions. In particular, the multiple index model allows for complex interactions but non-linearity is weak because we are only using a few projections whereas in CAM, we can capture the non-linearity but interactions are completely ignored, and the quadratic model assumes a linear structure with interactions without any provision for non-linearity. In general, these models assume a very particular structure, which limits the kind of non-linear patterns that can be learned, especially in terms of interactions within the time points for the predictors. Most of these approaches have been discussed in the review papers by \cite{articler1} and \cite{articler1m}.  


Neural Network (NN) and Deep Learning (DL) have attracted a great deal of attention and research due to their application in areas like time series \citep{Dorffner96neuralnetworks, qin2017dualstage, KHASHEI2010479}, image processing/computer vision \citep{gregor2015draw, Shi_2016_CVPR, doi:10.1080/014311697218719,Wang2019Development,71777188}, speech processing/natural language processing (NLP) \citep{inproceedings123, 6638947, 7472621,MORCHID201848}, genetics \citep{Singh2,Ehsan}, self-driving cars \citep{Maqueda_2018_CVPR,7995975}, and video learning \citep{yue2015beyond, karpathy2014large}. Neural network models by combining multiple linear combinations of the predictors after passing them through activation/link functions (called hidden layers) to learn complex non-linear relationships. Even with so much success, neural network and deep learning are still relatively undeveloped for functional data.  A traditional neural network includes no special provision for the temporal dependency seen in FDA. Recurrent Neural Network (RNN), especially Long Short-Term Memory (LSTM), have an edge over conventional feed-forward neural networks when it comes to time series or sequential data. This is due to their ability to selectively remember patterns for a long duration of time compared to regular RNN \citep{joshi2020machine}. The main drawback of deep learning techniques is that they can't handle curves directly, especially if the curves are measured at different time points or with a different number of observations, that is if the data is irregular.

\cite{1007599, articler4,ConanGuez2002MultilayerPF} developed and demonstrated the efficacy of Functional Neural Network (FNN). FNN was then further investigated by \cite{wang2019multilayer,wang2019remaining}, where they demonstrated the superiority of FNN to state-of-the-art alternatives (CNN, RNN, LSTM). The extension of the neural network to functional data is intriguing, 
as it enables efficient non-linear learning when the predictors are curves or surfaces. \cite{1007599} proposed a way to feed functional data as an input into a neural network and proved the validity by theoretical arguments and simulation experiments. But their initial work \citep{1007599} required the data to be dense and regular. This requirement was overcome with a slightly different FNN version \citep{ConanGuez2002MultilayerPF} where they used curve projections. \cite{wang2019multilayer} extended this work to deal with sparsity. The basic approach is that it only requires the entire time interval to be well covered by pooled observations across all the subjects. They pool measurements among all the subjects to estimate a set of basis functions and then use these basis functions to build a sparse functional neuron that extracts features for each subject. This method is developed for sparse FDA but it is equivalent to the original FNN with dense data. The main idea of FNN is that only the first layer is a functional hidden layer with functional neurons that give a scalar output, which is then connected to regular hidden layers and finally the output layer. In short, the functional layer is followed by a traditional neural network. This is a straight forward approach but it does not embrace the functional data completely. Other extensions of FNN for Spatio-Temporal modeling and Function-on-Function regression were explored in \cite{9260006,wang2020nonlinear}. We use the idea of FNN as an inspiration to develop our novel approach where we extend the initial functional hidden layer with functional neurons to multiple continuous hidden layers with continuous neurons with no regular hidden layers.

In this work, we study the problem of non-linear modeling from functional data using neural networks. There are mainly two challenges: 1) How to design a continuous hidden layer consisting of continuous neurons for a neural network that can learn functional representation from the functional inputs; and 2) how to train such a model. We propose a novel approach that takes the functional input and gives a scalar output, which can be solved using either Functional Direct Neural Network (FDNN) or Functional Basis Neural Network (FBNN). The main contributions of this work are as follows:
\begin{itemize}
\setlength\itemsep{0in}
    \item We introduce a new structure for non-linear modeling of functional data with a scalar response.
    \item We give two strategies, FDNN and FBNN, to solve the problem.
    \item We derive the functional gradient based learning algorithm for both strategies to optimize the parameters.
    \item We conduct simulation studies to demonstrate the effectiveness of FDNN and FBNN in varying settings. We also apply the models to several real-world examples.
\end{itemize}

The rest of this paper is organized as follows: In Section 2, we first present some background on FDA and FNN. Then we describe our approach and give two strategies to solve it, Functional Direct Neural Network (FDNN) and Functional Basis Neural Network (FBNN).  We follow this with simulations and several benchmark applications in Section 3. We consider multiple simulation settings depending on how the non-linear relation is defined between the response and functional predictor.  We demonstrate the efficiency of our methods over current state-of-the-art approaches. In the final section, we present our concluding remarks and future research directions which pertain to deeper statistical theory and extension to a variety of functional data settings.

\section{Methods}

\label{sec:meth}

We begin by providing the necessary notation used in the paper, followed by a summary of a few of the major non-linear models for scalar-on-function regression before describing our methodology. Suppose we have $N \in \mathcal{N}$ subjects observed over a compact time interval $\mathcal{T}$, and for the $i^{th}$ subject, the inputs are $R$ random functions that can be denoted as \{$X_{i, 1}, \ldots, X_{i, R}\} \in \mathcal{L}^2(\mathcal{T})$. The output is a scalar variable $Y_{i} \in \mathcal{R}$ for $i=1, \ldots, N$. We assume $X_{i,r}:=\{X_{i,r}(t): t\in [0,1]\}$ for $r=1, \ldots, R$  are fully observed functions, which is also known as dense functional data analysis \citep{FDA}. The main objective is to learn the non-linear mapping, $F:\mathcal{L}^2(\mathcal{T}) \times \dots \times \mathcal{L}^2(\mathcal{T}) \to \mathcal{R}$ from the functional inputs $X_{i,r}$ to the scalar output $Y_{i}$

\begin{equation} \label{eq 1}
\text{E}[Y_{i}|X_{i, 1}, \ldots, X_{i, R} ]=F\left(X_{i, 1}, \ldots, X_{i, R}\right).
\end{equation}

\subsection{Functional Regression}

There is vast literature in FDA when it comes to scalar-on-function regression and we briefly describe the most popular ones, which will be later used in simulations for comparison of different methods. In these models, the integrals without limits are taken over the entire domain.

The scalar-on-function regression model under different relation between the functional predictors and the scalar response is as follows:
\begin{itemize}
    \item Linear Model: 
    \begin{equation} \label{1}
    \text{E}[Y_{i}|X_{i, 1}, \ldots, X_{i, R} ]=\alpha+\sum_{r=1}^R\int \beta_r(t) X_{i,r}(t) dt
    \end{equation}
    where $\beta_r(t)$ is the parameter function \citep{FDA,ramsay1997functional}. This is also known as Funcitonal Linear Model (FLM).

    \item Continuously Additive Model (CAM): 
    \begin{equation} \label{2}
    \text{E}[Y_{i}|X_{i, 1}, \ldots, X_{i, R} ]=\sum_{r=1}^R\int f(X_{i,r}(t),t)dt 
    \end{equation}
    where bivariate parameter function $(x,t) \rightarrow f(x,t)$ is smooth \citep{McLean,articlel1,articlel2,articlel3,articlel4}.

    \item Single Index Model:
    \begin{equation}\label{3}
    \text{E}[Y_{i}|X_{i, 1}, \ldots, X_{i, R} ]=g\left(\sum_{r=1}^R <\beta_{r},X_{i,r}>\right) 
    \end{equation}
    where the function $g()$ is any smooth function defined on the real line \citep{EILERS2009196,articlem2}.
    
    \item Multiple Index Model:
    \begin{equation}\label{4}
    \text{E}[Y_{i}|X_{i, 1}, \ldots, X_{i, R} ]=g\left(\sum_{r=1}^R <\beta_{1r},X_{i,r}>,...,\sum_{r=1}^R<\beta_{Pr},X_{i,r}>\right) 
    \end{equation}
    Multiple Index models are extensions of the single index models \citep{articlem2,articlem3,article12,article13}.
    
    \item Quadratic Model:
    \begin{align} \label{5}
    \text{E}[Y_{i}|X_{i, 1}, \ldots, X_{i, R} ] & =\sum_{r=1}^R\int \beta_r(t) X_{i,r}(t) dt+\sum_{r=1}^R\int \int \beta(s,t) X_{i,r}(t)X_{i,r}(s) dtds, \\ 
     \label{6}
    \text{E}[Y_{i}|X_{i, 1}, \ldots, X_{i, R} ] &= \sum_{r=1}^R\int f(X_{i,r}(t),t)dt +\sum_{r=1}^R\int \int f(X_{i,r}(t),X_{i,r}(s),s,t)dtds 
    \end{align}
    Equation \eqref{5} represents a Quadratic Model given by \cite{quad} whereas Equation \eqref{6} represents a  Complex Quadratic Model defined for simulation purposes.
    
\end{itemize}


    
    
    

The above models are the most common approaches used in functional regression for linear and non-linear cases. They are commonly fit using basis expansions, Functional Principal Component Analysis (FPCA), or penalization methods. The non-linear models are analogous to their multivariate counterparts, which help to overcome the curse of dimensionality. In practice, it is difficult to know the actual form of the relation between the response and the functional predictors. Therefore, having a method that can deal with any kind of relationship without knowing the truth will be very useful.

\subsection{Functional Neural Network}


A neural network consists of three components: an input layer, hidden layers, and an output layer.  The input layer and output layer consist of neurons/nodes that are linked by multiple hidden layers consisting of neurons. In the forward propagation, that is to construct a predicted value, each neuron sums the weighted signal it receives from the input connections and produces an output signal which is a non-linear function of these inputs. Back propagation \citep{bp} (that is gradient descent) is used to determine the correct weights that minimize the overall cost in the neural network.

\begin{figure}[]
\centering
\includegraphics[height=5cm]{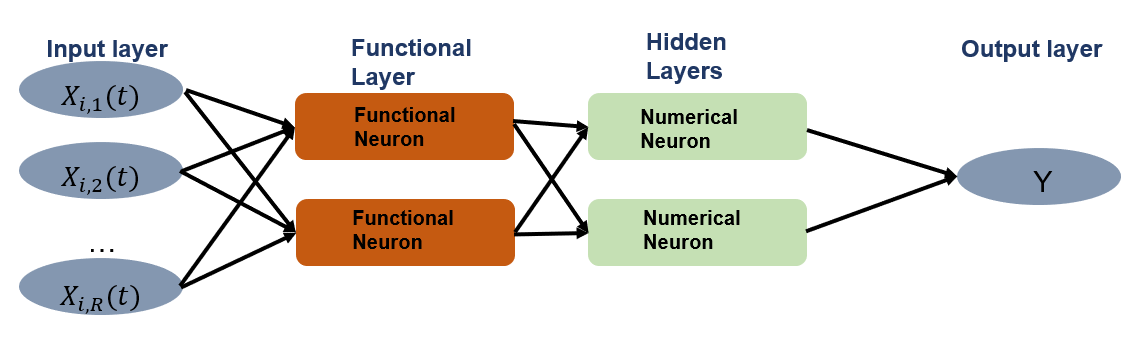}
  \caption{The architecture of a FNN with two functional neurons in the first layer and two numerical neurons in the second layer.}
  \label{FNN1}
\end{figure}

Figure \ref{FNN1} shows the architecture of the FNN model developed by \cite{1007599} using the idea of a generalized neuron \citep{STINCHCOMBE1999467}.
In FNN, the first hidden layer is a functional hidden layer which consists of functional neurons that give scalar outputs. These scalar values are then fed into a regular neural network framework. A functional neuron is defined as follows:
\begin{equation} \label{e1}
H(X,W_r)=\sigma \Big(b + \sum_{r=1}^{R} \int W_{r}(t)X_{r}(t)dt \Big), 
\end{equation}
where $b \in \mathcal{R}$ is the unknown intercept, $W_{r}\left( t\right)$ is the parameter function ($r=1, \ldots, R$) and $\sigma(\cdot)$ is a non-linear activation function from $\mathcal{R}$ to $\mathcal{R}$.

In forward propagation, as seen in Figure \ref{FNN1}, we first pass through the functional layer having functional neurons. For the value of $H(X,W_r(t))$, we use Equation \eqref{e1}, where numerical approximation is used to get the integral value. The forward propagation calculation in the subsequent hidden layers is straightforward because of the regular neural network layers. In the backpropagation step, we can go from the output layer to the second hidden layer using the partial derivatives that can be easily computed. It is critical to make sure that the partial derivatives at the first hidden layer for the parameters $W_r(t)$ exist. Under certain assumptions \citep{1007599, wang2019multilayer}, we can compute the partial derivatives required in the first hidden layer as:
\begin{equation} \label{e2}
\frac{\partial H(X,W_r)}{\partial W_{r}}(t)=\sigma^\prime \Big(b + \sum_{r^\prime=1}^{R} \int W_{r^\prime}(t^\prime)X_{r^\prime}(t^\prime)dt^\prime \Big)\times X_{r}(t)
\end{equation}
where $\sigma^\prime(\cdot)$ represents the first derivative of the activation function $\sigma(\cdot)$.

\subsection{Functional Direct Neural Network and Functional Basis Neural Network}

In our proposed approach, we develop a continuous mapping from layer to layer. To do this, we make use of what we call \textit{continuous neurons} and refer to a hidden layer with multiple continuous neurons as a \textit{continuous hidden layer}. In our neural network framework, we have three types of layers: an input layer, which is the first layer that takes in the functional predictors, the continuous hidden layers containing continuous neurons, and a functional output layer that provides the final fitted value using functional neurons. Unlike neural network, which learns using scalar weights, we learn using weight functions that are continuous over time (or some other continuum). The $l^{th}$ continuous hidden layer and its $k^{th}$ continuous neuron is defined as:
\begin{equation}
\begin{split}
H^{(l)}_{(k)}(s)&=\sigma \Big(b^{(l)}_{(k)}(s) + \sum_{j=1}^{J} \int w^{(l)}_{(j,k)}(s,t)H^{(l-1)}_{(j)}(t)dt \Big)\\
\widehat{Y}=H^{(L+1)}&=\sigma \Big(b^{(L+1)} + \sum_{j=1}^{J} \int w^{(L+1)}_{(j)}(t) H^{(L)}_{(j)}(t)dt \Big) \label{e3}
\end{split}
\end{equation}
where $l=1,2,3,...,L$, $H^{(0)}_{(j)}(t):=X_j(t)$, $b^{(L+1)} \in \mathcal{R}$ is the unknown intercept (also called the bias in machine learning), $b^{(l)}_{(k)}(s) \in \mathcal{L}^2(T)$ is the unknown intercept function, $w^{(l)}_{(j,k)}(s,t)$ is the bivariate parameter function for the $k^{th}$ continuous neuron in the $l^{th}$ hidden layer coming from the $j^{th}$ continuous neuron of the $(l-1)^{th}$ hidden layer, $w^{(L+1)}_{(j)}(t)$ is a parameter function for output layer coming from the $j^{th}$ continuous neuron of the $(L)^{th}$ hidden layer and $\sigma(\cdot)$ is a non-linear activation function. 

\begin{figure}[]
\centering
\includegraphics[height=4.5cm]{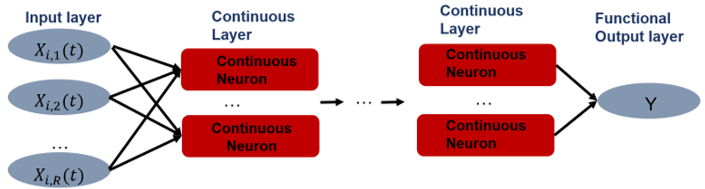}
  \caption{General architecture for our proposed approach with multiple continuous hidden layers.}
  \label{FDNN2}
\end{figure}

Comparing Figure \ref{FNN1} to Figure \ref{FDNN2}, and Equation \eqref{e1} to Equation \eqref{e3}, we can clearly see the difference between our approach and the functional neural network defined by \cite{1007599}. Our approach uses continuous hidden layers consisting of continuous neurons that output a function and learn bivariate functional weights, unlike the FNN where only the first layer uses functional neurons that output a scalar value. We use the functional neurons only in the last layer to get the final output value. Also, unlike FNN where the functional layer is connected to multiple regular hidden layers of classical NN, our approach has multiple continuous hidden layers connected to each other and a functional layer at the end, which gives the scalar output value. 
A major motivation for this approach is to preserve the functional nature of the data for as long as possible to provide a richer structure for the prediction that also exploits the domain information and continuity of the predictor functions.

We implement our proposed approach using two strategies: (1) a more direct solution when the data is observed on a common grid or (2) using basis expansions to allow for more irregular sampling of the functional predictors. The former approach we call as Functional Direct Neural Network (FDNN) while the latter approach we refer to as Functional Basis Neural Network (FBNN). FDNN works well with densely observed data on a common grid. We can apply standard tools from the neural network in FDNN directly, making it straightforward to implement. We are learning the weight functions similar to the way we learn the weights in the neural network and also have the flexibility to change the grid length (s) between the continuous hidden layers. FBNN uses basis expansion to learn the relation and also allows for greater parsimony if the data is sampled on a very dense grid, as we need to only learn the weights of the basis expansion. 


\textbf{Functional Direct Neural Network (FDNN)}: For simplicity, we assume that Figure \ref{FDNN2} has $R$ inputs (functional predictors), $L$ continuous hidden layers, each with $J$ incoming and $K$ outgoing connections. FDNN has the same flexibility as a neural network, where we can specify the number of continuous hidden layers and the number of continuous neurons in each of them. Commonly used activation functions in the continuous neurons include ReLU, tanh, sigmoid/logistic and linear. The activation function used in the functional output layer depends on the response. In our simulations, for the continuous response, we use the linear activation function in the functional output layer, whereas for the binary response, we use the sigmoid activation function. 
In the forward propagation phase, we reach from the functional predictors to the scalar response via a chain of multiple CAMs, and in the backpropagation phase, we want to learn the weight functions and the intercept functions.

We derive a functional gradient based optimization algorithm to optimize the network parameters. The gradient of the weight function measures the change in a functional to a change in a function on which the functional depends. The needed assumptions and mathematical tools can be found in \cite{1007599, wang2019multilayer}. We extend the gradient descent optimization technique to the functional setting using calculus of variations \citep{booko}, in particular by taking Fr\'echet derivatives. In the backpropagation step, we pass through the network in reverse, as we do in a neural network, but we calculate the partial derivative for the weight functions, the intercept functions and the intercept. Let the loss function be $\mathcal{L}(\theta)$, 
where $\theta$  is the collection of all functional and scalar parameters. The loss function is defined based on the response, quadratic loss for a continuous response and a cross entropy loss for a binary response. 
The partial derivatives required for the parameters of the continuous neurons in the continuous hidden layers of our proposed architecture (as shown in Figure \ref{FDNN2}) are as follows:
\begin{equation} \label{e5}
\begin{split}
\frac{\partial H^{(l)}_{(k)}}{\partial b^{(l)}_{(k)}}(s)&= \sigma^\prime \Big(b^{(l)}_{(k)}(s) + \sum_{j=1}^{J} \int w^{(l)}_{(j,k)}(s,t)H^{(l-1)}_{(j)}(t)dt \Big)\\
\frac{\partial H^{(l)}_{(k)}}{\partial w^{(l)}_{(j,k)}}(s,t)&=\sigma^\prime \Big(b^{(l)}_{(k)}(s) + \sum_{j^\prime=1}^{J} \int w^{(l)}_{(j^\prime,k)}(s,t^\prime)H^{(l-1)}_{(j^\prime)}(t^\prime)dt^\prime \Big) \times H^{(l-1)}_{(j)}(t)
\end{split}
\end{equation}
where  $l=1,2,3,...,L$, $H^{(0)}_{(j)}(t):=X_j(t)$, and $\sigma^\prime(\cdot)$ represents the first derivative of $\sigma(\cdot)$.

Next, the partial derivatives required for the parameters of the functional neurons in the functional output layer are,
\begin{equation} \label{e4}
\begin{split}
\frac{\partial H^{(L+1)}} {\partial b^{(L+1)}}&=\sigma^\prime \Big(b^{(L+1)} + \sum_{j=1}^{J} \int w^{(L+1)}_{(j)}(t) H^{(L)}_{(j)}(t)dt \Big)\\
\frac{\partial H^{(L+1)}} {\partial w^{(L+1)}_{(j)}}(t)&=\sigma^\prime \Big(b^{(L+1)} + \sum_{j^\prime=1}^{J} \int w^{(L+1)}_{(j^\prime)}(t^\prime) H^{(L)}_{(j^\prime)}(t^\prime)dt^\prime \Big) \times H^{(L)}_{(j)}(t) 
\end{split}
\end{equation}
Appendix A contains in-depth information regarding the partial derivatives. We can consider the values of the grid/domain ($s$), the number of continuous hidden layers, and the number of continuous neurons in each continuous hidden layer as hyperparameters. The general way of optimizing for the parameters is still the same in FDNN, using backpropagation with the help of the gradients given above.

\textbf{Functional Basis Neural Network (FBNN)}: Similar to FDNN, we have continuous hidden layers with continuous neurons in FBNN, but the continuous neurons are expanded using basis expansion. Unlike FDNN, we learn the weights for the basis functions rather than the weight function itself. As we are using basis functions, we can deal with irregular data more easily \citep{FDA}. We get the forward feed in FBNN by using basis expansion where we replace the weight functions ($w^{(l)}_{(j,k)}(s,t)$) in Equation (\ref{e3}) by a linear combination of basis functions as seen below. We will again see the details of our approach using the architecture shown in Figure \ref{FDNN2}. Let subscripts $j,k$ represent the continuous neurons of the neural network and superscript $l$ denote the $l^{th}$ layer for the weight parameter $(W^{(l)}_{(j,k)})$. We define $v$ to be some standard collection of basis functions, like splines, wavelets, or sine and cosine functions. Also, let us assume that we use the same basis function and the same number of basis in each of the continuous neurons in the continuous hidden layer across the network for simplicity. This gives us the following Equation:
\begin{equation} \label{e8}
\begin{split}
H^{(l)}_{(k)}(s)&=\sigma \Big(b^{(l)}_{(k)}(s) + \sum_{j=1}^{J} \int w^{(l)}_{(j,k)}(s,t)H^{(l-1)}_{(j)}(t)dt \Big)\\
&=\sigma \Big(\sum_{b=1}^{B} b^{(l)}_{(k),b} v^{*}_b(s)  + \sum_{j=1}^{J}  \sum_{c=1}^{C} \sum_{d=1}^{D}  w^{(l)}_{(j,k),c,d}  v_c(s)  \int v_d(t) H^{(l-1)}_{(j)}(t)dt \Big)\\
&=\sigma \Big(\sum_{b=1}^{B} b^{(l)}_{(k),b} v^{*}_b(s)  + \sum_{j=1}^{J}  \sum_{c=1}^{C} \sum_{d=1}^{D}  w^{(l)}_{(j,k)c,d}  v_c(s)  A_{(j),d}^{(l)}\Big)\\
&=\sigma \Big(B^{(l)}_{(k)} v^{*}(s)  + \sum_{j=1}^{J}   W^{(l)}_{(j,k)}  v(s)  A_{(j)}^{(l)}\Big) \\
\end{split}
\end{equation}

\begin{equation} \label{e81}
\begin{split}
\widehat{Y}=H^{(L+1)}&=\sigma \Big(b^{(L+1)} + \sum_{j=1}^{J} \int w^{(L+1)}_{(j)}(t) H^{(L)}_{(j)}(t)dt \Big)\\
&=\sigma \Big(b^{(L+1)} + \sum_{j=1}^{J}  \sum_{c=1}^{C} w^{(L+1)}_{(j),c} \int v_c(t)  H^{(L)}_{(j)}(t)dt \Big)\\
&=\sigma \Big(b^{(L+1)} + \sum_{j=1}^{J}  \sum_{c=1}^{C} w^{(L+1)}_{(j),c}  A_{(j),c}^{(L+1)}\Big)=\sigma \Big(b^{(L+1)} + \sum_{j=1}^{J}  W^{(L+1)}_{(j)}  A_{(j)}^{(L+1)}\Big)
\end{split}
\end{equation}

where $l=1,2,3,...,L$, $H^{(0)}_{(j)}(t):=X_j(t)$, and $b^{(L+1)} \in \mathcal{R}$ is the unknown intercept. $B^{(l)}_{(k)}$, $W^{(l)}_{(j,k)}$ and $W^{(L+1)}_{(j)}$ are the unknown basis weights corresponding to their respective basis functions $v^{*}(s),$ $v(s),$ and $v(t)$. $B$, $C$ and $D$ specify the number of basis used in each of the continuous neuron of the continuous hidden layers. $\sigma(\cdot)$ is a non-linear activation function. 

For FBNN, we can again extend to as many continuous hidden layers as we want with any number of continuous neurons. The activation function selection is the same as FDNN.  We can consider values of grid/domain ($s$), the number of continuous hidden layers, the number of continuous neurons in each continuous hidden layer, the selection of the basis functions, and the number of basis as hyperparameters. We learn using the basis weights of the basis functions, to reach from the functional predictors to the scalar response via a chain of multiple CAMs in the forward propagation phase. The partial derivatives required for the parameters of the continuous neurons in the continuous hidden layers under FBNN in our proposed architecture (as seen in Figure 2) are as follows:
\begin{equation} \label{e10}
\begin{split}
\frac{\partial H^{(l)}_{(j)}(s)}{\partial B^{(l)}_{(k)}}&= \sigma^\prime \Big(B^{(l)}_{(k)}v^{*}(s) + \sum_{j=1}^{J}   W^{(l)}_{(j,k)}  v(s)  A_{(j)}^{(l)}\Big)\times v^{*}(s)\\
\frac{\partial H^{(l)}_{(j)}(s)}{\partial W^{(l)}_{(j,k)}}
&= \sigma^\prime \Big(B^{(l)}_{(k)}v^{*}(s) + \sum_{j^\prime=1}^{J}   W^{(l)}_{(j^\prime,k)}  v(s)  A_{(j^\prime)}^{(l)}\Big) \times  v(s)  A_{(j)}^{(l)} 
\end{split}
\end{equation}
where $l=1,2,3,...,L$, $H^{(0)}_{(j)}(t):=X_j(t)$, and $\sigma^\prime(\cdot)$ represents the first derivative of $\sigma(\cdot)$.

Next, the partial derivatives required for the parameters of the functional neurons in the functional output layer are,
\begin{equation} \label{e9}
\begin{split}
\frac{\partial H^{(L+1)}} {\partial b^{(L+1)}}&=\sigma^\prime \Big(b^{(L+1)} + \sum_{j=1}^{J}  W^{(L+1)}_{(j)}  A_{(j)}^{(L+1)}\Big) \\
\frac{\partial H^{(L+1)}} {\partial W^{(L+1)}_{(j)}}&=\sigma^\prime \Big(b^{(L+1)} + \sum_{j^\prime=1}^{J}  W^{(L+1)}_{(j^\prime)}  A_{(j^\prime)}^{(L+1)}\Big) \times  A_{(j)}^{(L+1)}
\end{split}
\end{equation}

The general way of optimizing for the parameters is the same in FBNN, using backpropagation, but the gradients are found using the above partial derivatives.

\textbf{Regularization}: Over-fitting is a common problem with machine learning and nonparametric statistical methods.  They are often over-parameterized and thus, care must be taken to ensure that the results still produce reliable out of sample predictions. Regularization in neural network to deal with over-fitting is an active area of research \citep{Yang2020ProxSGD, LUDWIG201433, 8603826}. Common methods used to address this issue include penalization, drop out, data augmentation, and early stopping method. We concentrate on early stopping, which utilizes a validation set to stop the model fitting process early before it starts over-fitting. Naturally, both FDNN and FBNN tend to over-fit if the models are too deep or run for a large number of iterations. Early stopping is easy to implement and, as we will see in Section 3, it is effective at improving the performance of FDNN and FBNN.

\section{Empirical Results}
\label{sec:verify}

In this section, we present multiple simulations to evaluate the effectiveness of 
FDNN and FBNN, as well as benchmarking with multiple real world data sets. We demonstrate the following objectives through our results: 1) FDNN and FBNN are effectively trained by the derived optimization method 2) FDNN and FBNN capture the relation between the functional predictors and the scalar response 3) FBNN and FDNN outperform compared to other methods.

\subsection{Simulation Study}

For simplicity, we consider a single predictor function (that is R=1) which is dense and regular. We simulate $n=1500$ iid random curves $\left\{X_{1}(t), \cdots, X_{n}(t)\right\}$ from a Gaussian process with mean 0 and covariance
$$
  C_{X}(t, s)=\frac{\sigma^{2}}{\Gamma(\nu) 2^{\nu-1}}\left(\frac{\sqrt{2 \nu}|t-s|}{\rho}\right)^{\nu} K_{\nu}\left(\frac{\sqrt{2 \nu}|t-s|}{\rho}\right),
$$
which is the Matérn covariance function and $K_{\nu}$ is the modified Bessel function of the second kind. We set $\rho=0.5,$ $\nu=5 / 2$ and $\sigma^{2}=1 .$ These curves are evaluated at $m=200$ equally-spaced time points from $[0,1]$. 

The data generating models for the continuous and binary response are given in Tables \ref{ss1} and \ref{ss2} respectively. We add $\epsilon_i$ (random noise) to our continuous response model to adjust the signal to noise ratio. For the binary response, we assume that we have an outcome $Y_{i}$ related to function $X_{i}$ via a link function $g:$ $ E(Y_i|X_i)=g^{-1}\left(\eta_{i}\right)$ where $\eta_{i}$ is defined as the response in any of the models mentioned in Section (2.2). Since we have a binary response, the link function used is the logistic function, which is also the canonical link function. Each of these settings is simulated 100 times and we divide the 1500 samples 
into 1000 for training and 500 for testing. For the Deep Learning methods, we use early stopping which requires a validation set. We use 500 samples from the training set as the validation set in such cases. We are primarily interested in the prediction performance and therefore, we report the Root Mean Square Error (RMSE) of the prediction for the continuous response, and the classification error as well as the log-likelihood for the binary response.

For FDNN and FBNN, we consider multiple combinations of continuous hidden layers $(l=1,2)$ and continuous neurons $(j=1,2,4)$. The value for the grid $(s)$ in the continuous neurons is in the range $(30,300)$. For FBNN, we use B-splines as the basis functions and for the knots, we consider equal spacing. We tune the number of B-splines which corresponds to the number of inner knots, for which consider the range $(5,10)$. We determine the best values of the hyper parameters using grid search. We observe that mostly the number of continuous neurons and the number of continuous hidden layers change the result drastically while other hyper-parameters have marginal effect.

\begin{table}[]
\centering
\footnotesize
\begin{tabular}{l} 
Simulation Scenarios \\
\hline 1. Linear:\text{ }$Y_{i}=\alpha+\sum_{r=1}^R\int \beta_r(t) X_{i,r}(t) dt+\epsilon_i \qquad\quad \text{ where } \beta(t)=5 \times \sin (2 \pi t) \text { and } \alpha=0$\\
2. CAM:\text{ }$Y_i=\sum_{r=1}^R\int f(X_{i,r}(t),t)dt + \epsilon_i \qquad \quad\quad \hspace{3.9mm} \text { where } f(X_{i,r}(t),t)=(X_{i,r}(t)^2)$\\
3. Single Index:$\text{ }Y_i=g\left(\sum_{r=1}^R <\beta_{r},X_{i,r}>\right) + \epsilon_i \quad \hspace{2.2mm} \text { where } g(a)=a^2, \text { } \beta(t)=5 \times \sin (2 \pi t)$\\ 
4. Multiple Index:$\text{ }Y_i=g\left(\sum_{r=1}^R <\beta_{1,r},X_{i,r}>,...,\sum_{r=1}^R<\beta_{P,r},X_{i,r}>\right) + \epsilon_i \quad$\\
$\qquad\qquad\qquad\qquad\qquad\qquad\qquad\qquad\qquad\qquad\quad  \hspace{1.mm}\text { where } g(a,b)=a^2+b^2, \text { } \beta_1(t)=5 \times \sin (2 \pi t)$,\\
$\qquad\qquad\qquad\qquad\qquad\qquad\qquad\qquad\qquad\qquad\qquad\quad\quad \hspace{1.25mm}  \beta_2(t)=5 \times \sin (3 \pi t)$\\ 
5. Quadratic:$\text{ }Y_i=\sum_{r=1}^R\int \beta_r(t) X_{i,r}(t) dt+\sum_{r=1}^R\int \int \beta(s,t) X_{i,r}(t)X_{i,r}(s) dtds + \epsilon_i \qquad\qquad$\\
$\qquad\qquad\qquad\qquad\qquad\qquad\qquad\qquad\qquad\qquad\quad \hspace{1.mm}\text { where } \beta(t,s)=\beta_1(t)*\beta_2(s),\text { } \beta_1(t)=5 \times \sin (3 \pi t),$\\
$\qquad\qquad\qquad\qquad\qquad\qquad\qquad\qquad\qquad\qquad\qquad\quad\quad\hspace{1.5mm}\beta_2(s)=5 \times \sin ( \pi s).$\\ 
6. Complex Quadratic:$\text{ }Y_i=\sum_{r=1}^R\int f(X_{i,r}(t),t)dt +\sum_{r=1}^R\int \int f(X_{i,r}(t),X_{i,r}(s),s,t)dtds + \epsilon_i \qquad\qquad$\\
$ \qquad\qquad\qquad\qquad\qquad\qquad\qquad\qquad\qquad\qquad\quad\hspace{1.mm}\text  { where } f(X_{i,r}(t),t)=(X_{i,r}(t)^2),$\\ $\qquad\qquad\qquad\qquad\qquad\qquad\qquad\qquad\qquad\qquad\qquad\quad\quad\hspace{1.5mm}f(X_{i,r}(t),X_{i,r}(s),s,t)=(X_{i,r}(t)\times X_{i,r}(s))^2$\\
\hline

\end{tabular}\\
\caption{Different simulation scenarios with continuous response and a single predictor function.}
\label{ss1}
\end{table}

In Table \ref{t0}, we summarize the performance of each method under different simulation settings with a continuous response. 
We observe that when the true model is Linear, four models do equally well: FLM, CAM, and both of our methods. When the truth is CAM, then our approaches along with CAM do equally well. More substantial gains are realized as the true model becomes more complex and interactions are considered. In such situations, our methods dominate others. Interestingly, under the Complex Quadratic settings, all the methods struggle substantially compared to our approach, but there is still  potential for improvement (say, through deeper networks) as the optimal RMSE is one (since that is the variance of the error). The gain from early stopping can be seen in Appendix B. Table \ref{tt2} shows a similar trend with the binary response. When the truth is Logistic or CAM, then FLM, CAM, and both of our methods perform equally well. Our methods outperform others as the true model becomes more complex and interactions are considered. We see similar results when comparing the log-likelihood values as seen in Appendix C. In general, we see superior results to other approaches in several settings, and we are always comparable if not better to the best method in each setting. Note that all the table values for the deep learning approaches are obtained using early stopping. We have also compared the deep learning approaches under the same parameter setting in Appendix C.

\begin{table}[]

\centering
\begin{tabular}{|l|l|l|l|l|l|l|l|}
\hline
Mapping & FLM            & CAM            & NN & FNN  & FDNN           & FBNN                      \\ \hline
Linear          & {1.011} & {1.015} & 1.185    & 1.150  & {1.019} & {1.011}               \\ \hline
CAM          & 1.705          & {1.053} & 1.451    & 1.298 & {1.076} & {1.058}                   \\ \hline
Single Index           & 1.711           & 1.532          & 1.515    & 1.225 & {1.067} & {1.061}          \\ \hline
Multiple Index           & 1.808          & 1.350          & 1.609    & 1.290 & {1.100} & {1.060}   \\ \hline
Quadratic          & 1.646          & 1.211          & 1.480    & 1.273 & {1.092} & {1.092}      \\ \hline
Complex Quadratic          & 1.807          & 1.414          & 1.905    & 1.632 & {1.232} & {1.202}  \\ \hline
\end{tabular}
\caption{Comparison of the RMSE for the methods under different setting between the functional predictors and the continuous response.}
\label{t0}
\end{table}

\begin{table}[]
\centering
\footnotesize

\begin{tabular}{l} 
Simulation Scenarios \\
\hline 1. Logistic:$\text{ logit}(\eta_i)=\alpha+\sum_{r=1}^R\int \beta_r(t) X_{i,r}(t) dt \qquad\quad\quad \text{ where } \beta(t)=5 \times \sin (2 \pi t) \text { and } \alpha=0$\\
2. CAM:$\text{ logit}(\eta_i)=\sum_{r=1}^R\int f(X_{i,r}(t),t)dt  \qquad \quad\quad\quad \hspace{5.5mm} \text { where } f(X_{i,r}(t),t)=sin(X_{i,r}(t))$\\
3. Single Index:$\text{ logit}(\eta_i)=g\left(\sum_{r=1}^R <\beta_{r},X_{i,r}>\right)  \quad\quad \hspace{3.5mm}  \text { where } g(a)=sin(a),\text { } \beta(t)=5 \times \sin (2 \pi t)$\\ 
4. Multiple Index:$\text{ logit}(\eta_i)=g\left(\sum_{r=1}^R <\beta_{1,r},X_{i,r}>,...,\sum_{r=1}^R<\beta_{P,r},X_{i,r}>\right)  \quad$\\
$\qquad\qquad\qquad\qquad\qquad\qquad\qquad\qquad\qquad\qquad\hspace{1.25mm}\text { where } g(a,b)=sin(sin(a)+b),\text { } \beta_1(t)=5 \times \sin (2 \pi t),$\\
$\qquad\qquad\qquad\qquad\qquad\qquad\qquad\qquad\qquad\qquad\qquad\quad\hspace{1.5mm}  \beta_2(t)=5 \times \sin (3 \pi t)$\\ 
5. Quadratic:$\text{ logit}(\eta_i)=\sum_{r=1}^R\int \beta_r(t) X_{i,r}(t) dt+\sum_{r=1}^R\int \int \beta(s,t) X_{i,r}(t)X_{i,r}(s) dtds  \qquad\qquad$\\
$\qquad\qquad\qquad\qquad\qquad\qquad\qquad\qquad\qquad\qquad\hspace{1.25mm}\text { where } \beta(t,s)=\beta_1(t)*\beta_2(s),\text { } \beta_1(t)=5 \times \sin (3 \pi t),$\\
$\qquad\qquad\qquad\qquad\qquad\qquad\qquad\qquad\qquad\qquad\qquad\quad\hspace{1.5mm}\beta_2(s)=5 \times \sin ( \pi s).$\\ 
6. Complex Quadratic:$\text{ logit}(\eta_i)=\sum_{r=1}^R\int f(X_{i,r}(t),t)dt +\sum_{r=1}^R\int \int f(X_{i,r}(t),X_{i,r}(s),s,t)dtds  \qquad\qquad$\\
$\qquad \qquad\qquad\qquad\qquad\qquad\qquad\qquad\qquad\qquad\hspace{1.25mm}\text  { where } f(X_{i,r}(t),t)=sin(X_{i,r}(t)),$\\ $\qquad\qquad\qquad\qquad\qquad\qquad\qquad\qquad\qquad\qquad\qquad\quad\hspace{1.5mm}f(X_{i,r}(t),X_{i,r}(s),s,t)=sin(X_{i,r}(t)\times X_{i,r}(s))$\\
\hline

\end{tabular}\\
\caption{Different simulation scenario with binary response and a single predictor function.}
\label{ss2}
\end{table}

\begin{table}[]
\centering
\begin{tabular}{|l|l|l|l|l|l|l|l|}
\hline
Mapping & FLM            & CAM            & NN     & FNN     & FDNN           & FBNN                       \\ \hline
Logistic          & {0.276} & {0.276} & 0.291 & 0.288    & {0.278} & {0.277}                \\ \hline
CAM          & {0.390}          & {0.392} & 0.402  & 0.398  & {0.393} & {0.392}                \\ \hline
Single Index          & 0.365          & 0.366          & 0.373 & 0.372   & {0.357} & {0.355}         \\ \hline
Multiple Index          & 0.374          & 0.376          & 0.392 & 0.410    & {0.370} & {0.368}        \\ \hline
Quadratic          & 0.363          & 0.328          & 0.344 & 0.338   & {0.322} & {0.324}    \\ \hline
Complex Quadratic          & {0.394}          & {0.392}          & 0.403 & 0.398    & {0.392} & {0.393}\\ \hline
\end{tabular}
\caption{Comparison of the classification error for the methods under different setting between the functional predictors and the binary response.}
\label{tt2}
\end{table}

\subsection{Real Data}

\begin{figure}[]
\centering
\includegraphics[height=10cm]{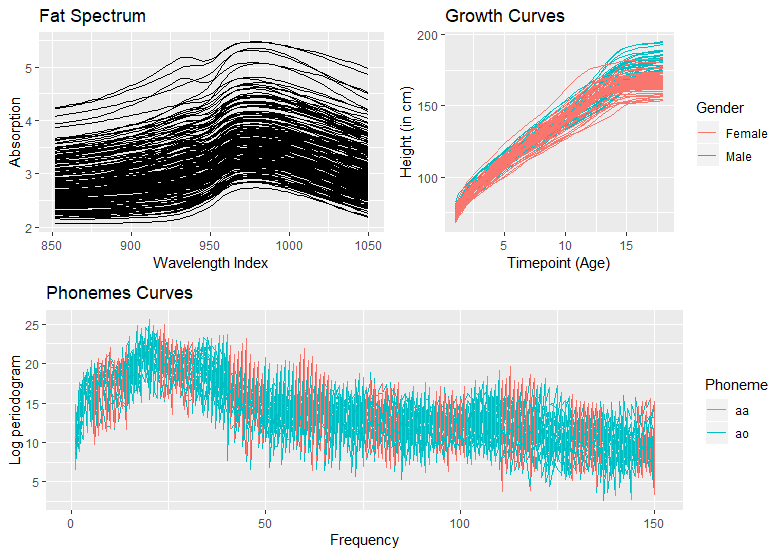}
  \caption{Sample curves for each of the three data set.}
  \label{real3}
\end{figure}

We analyze three different data sets as seen in Figure \ref{real3}. The fat content spectrometric data \citep{10} consists of food samples that contain finely chopped pure meat with different fat content, the growth curves from Berkeley Growth Study \citep{articlebg} contains information about the heights (in cm) of the children at different age and the speech recognition data from TIMIT (available at https://hastie.su.domains/ElemStatLearn/) where we have speech signals for different phonemes. The spectrometric data is recorded on a Tecator Infractec Food and Feed Analyzer working in the wavelength range 850-1050nm by the near-infrared (NIR) transmission principle. For each meat sample, the data consists of a 100 channel spectrum of absorbances (-log10 of the transmittance measured by a spectrometer) and a fat content measure derived from an analytic chemical processing. We predict the fat content of a meat sample on the basis of its NIR absorbance spectrum. The Growth data contains the heights of 39 boys and 54 girls from age 1 to 18 and the ages at which they were collected. The goal is to build a classifier for the gender using the growth curves. In the speech recognition data, the predictor is a voice signal as a function of time and the response is the name of the vowel sounds pronounced, which can be regarded as a discrete random variable. We only take a subset of the data, similar to \cite{li2017} and only consider two phonemes (response) transcribed as follows: “aa” as the vowel in “dark” and “ao” as the first vowel in “water”. The distinction between the two groups is not straightforward as seen in Figure \ref{real3}, which shows the phoneme curves for each of the two groups computed by a log-periodogram of 150 \citep{doi:10.1198/004017008000000154}. We use the speech signals as functional predictors to classify the different phonemes.

\begin{table}[]
\centering
\begin{tabular}{|l|l|l|l|l|l|l|}
\cline{1-7}
Data & FLM   & CAM   & NN    & FNN   & FDNN  & FBNN  \\ \cline{1-7}
Fat Spectrum  & 2.517 & 5.103 & 1.249 & 2.777 & {0.822} & {1.197}\\ \cline{1-7}
Growth   & 0.150 & 0.150 & 0.450 & 0.450 & {0.100} & {0.100}\\ \cline{1-7}
Phonemes   & 0.187 & 0.231 & 0.237 & 0.232 & {0.156} & {0.160}\\ \cline{1-7}
\end{tabular}
\caption{Comparison of different methods on the real data using RMSE (Fat Spectrum) and Classification Error (Growth and Phonemes).}
\label{t1}
\end{table}
In the Fat Spectrum data, we have 216 samples of which 176 are for training (40 samples for validation) and 40 samples are kept aside for testing. The response is the fat content value. It should be noted that the spectra are finely sampled, leading to very smooth curves as seen in Figure \ref{real3}. Next, we look into the growth curves from Berkeley Growth Study. The heights (in cm) of the children in the data set are measured from 1 to 18 years at 31 (m=31) timepoints. We divide the data into 73 and 20 samples for the training (20 samples for validation) and testing respectively. We take the response as Gender. In the last data set, we divide the data into 640 and 160 samples for training (160 samples for validation)  and testing respectively and the response is the phoneme. We can see from Table \ref{t1} that FDNN and FBNN are performing much better compared to all the traditional methods in the Spectrometric data for predicting the fat content of the meat sample. 
For the two classification problems, we see a similar trend where our methods have lower classification error compared to other methods whether the sample size is small or large. The true nature of the relation between the functional predictors and the response is never known in real world settings but our methods can learn any kinds of patterns and excel in such situations as seen in Table \ref{t1}.

\section{Discussion and Conclusions}

In this paper, a novel approach for complex functional regression with a scalar response has been developed. It is a natural extension of a neural network to functional data. We use continuous hidden layers with continuous neurons, allowing us to capture temporal dependency and complicated relationships. We have devised two ways to implement this approach, FDNN and FBNN. Our approach is flexible as we don't assume a particular structure that limits the kinds of non-linear patterns learned, especially, interactions within time points for the predictors.

Understanding both the methods further will help us determine when it is more beneficial to use FDNN or FBNN. Both, in general, seem to be working on par with the best existing methods under simple cases but are outperforming  in more complicated non-linear settings. We achieve superior results even with such shallow networks. The proposed approach is also appealing in practice as it provides a way for researchers to systematically conduct deep learning tasks when they have functional predictors. 
The main advantage of FDNN over FBNN is that it is straightforward to apply on dense and nicely gridded data, works on the same ideas as a neural network and requires less tuning, so it is a good approach to start with. FBNN is more flexible as it can deal with irregular samples but is dependent on the imputation quality as well as the selection of the basis functions. FBNN learns basis weights allowing for greater parsimony with very dense grids. Further theoretical investigation of these approaches is needed. Hence, an avenue for future work is developing statistical theory to give these methods sound backing. Also, the performance of FBNN in a sparse setting can be of interest but needs nontrivial extensions.

Lastly, Functional regression has been studied extensively (especially the scalar-on-function models) and non-linear modeling for functional data has received more attention recently. But there are still many questions that remain open, in particular function-on-function models. The applicability of such models is of huge significance as they can be used in many areas including time series, speech analysis, and computer vision. There is a need for a more sophisticated and flexible function-on-function model in such domains, as functional data becomes more prevalent in today’s world. Hence, extending our work to the function-on-function case will enable us to model functional predictors with functional response and help capture their complex relation.

\section*{Acknowledgement}
This research was supported in part by the following grant to Pennsylvania State University:  NSF SES-1853209

\bibliographystyle{apalike}
\bibliography{ref}

\newpage
\section{Appendix}

\section*{Appendix A. Mathematical Details for FDNN}
\label{app:theorem}



The partial derivatives used in FDNN are as follows,
\begin{equation} \label{a5}
\begin{split}
\frac{\partial H^{(l)}_{(k)}}{\partial w^{(l)}_{(j,k)}}(s,t) &=\sigma^\prime \Big(b^{(l)}_{(k)}(s) + \sum_{j^\prime=1}^{J} \int w^{(l)}_{(j^\prime,k)}(s,t^\prime)H^{(l-1)}_{(j^\prime)}(t^\prime)dt^\prime \Big)\\
&\times \int \frac{\partial}{\partial w^{(l)}_{(j,k)}(s,t)} w^{(l)}_{(j,k)}(s,t)H^{(l-1)}_{(j)}(t)dt \\
&=\sigma^\prime \Big(b^{(l)}_{(k)}(s) + \sum_{j^\prime=1}^{J} \int w^{(l)}_{(j^\prime,k)}(s,t^\prime)H^{(l-1)}_{(j^\prime)}(t^\prime)dt^\prime \Big) \times H^{(l-1)}_{(j)}(t)\\
\frac{\partial H^{(L+1)}} {\partial w^{(L+1)}_{(j)}}(t)&=\sigma^\prime \Big(b^{(L+1)} + \sum_{j^\prime=1}^{J} \int w^{(L+1)}_{(j^\prime)}(t^\prime) H^{(L)}_{(j^\prime)}(t^\prime)dt^\prime \Big)\\
&\times \int \frac{\partial }{\partial w^{(L+1)}_{(j)}(t)} w^{(L+1)}_{(j)}(t)H^{(L)}_{(j)}(t) dt\\ 
&=\sigma^\prime \Big(b^{(L+1)} + \sum_{j^\prime=1}^{J} \int w^{(L+1)}_{(j^\prime)}(t^\prime) H^{(L)}_{(j^\prime)}(t^\prime)dt^\prime \Big) \times H^{(L)}_{(j)}(t)
\end{split}
\end{equation}
where $\sigma^\prime(\cdot)$ represents the first derivative of the activation function $\sigma(\cdot)$.

\section*{Appendix B. Early Stopping and Log-likelihood} 

\begin{table}[h]
\centering
\footnotesize
\begin{tabular}{|l|l|l|l|l|l|l|l|}

\hline Mapping & FDNN & FBNN & FDNN (2,2) & FDNN (4,4) & FBNN (2,2) & FBNN (4,4)  \\
\hline Linear & ${1 . 0 6 9}$ & ${1 . 0 1 5}$ & 1.452 & 1.440 & 1.244 & 1.271  \\
\hline CAM & ${1 . 1 1 8}$ & ${1 . 0 8 4}$ & 1.460 & 1.459 & 1.217 & 1.202  \\
\hline Single Index & ${1 . 1 5 1}$ & ${1 . 3 5 9}$ & 1.457 & 1.461 & 1.243 & 1.200 \\
\hline Multiple Index & ${1 . 1 5 4}$ & ${1 . 0 9 1}$ & 1.495 & 1.473 & 1.207 & 1.219  \\
\hline Quadratic Linear & ${1 . 1 3 6}$ & ${1 . 2 0 5}$ & 1.495 & 1.465 & 1.284 & 1.263  \\
\hline Complex Quadratic & ${1 . 2 8 7}$ & ${1 . 3 4 5}$ & 1.586 & 1.591 & 1.386 & 1.309  \\
\hline
\end{tabular}
\caption{Comparison of Deep FDNN and FBNN under different settings between the functional predictors and the continuous response.}
\label{at1}
\end{table}

\begin{table}[h]
\centering
\footnotesize
\begin{tabular}{|l|l|l|l|l|l|l|l|}

\hline Mapping & FDNN & FBNN & FDNN (2,2) & FDNN (4,4) & FBNN (2,2) & FBNN (4,4)  \\
\hline Linear & ${1.019}$ & ${1.011}$ & 1.036 & 1.043 & 1.034 & 1.028  \\
\hline CAM & 1.076 & ${1.062}$ & 1.107 & 1.072 & 1.105 & ${1.058}$  \\
\hline Single Index & ${1.067}$ & 1.346 & 1.069 & 1.122 & 1.078 & ${1.061}$   \\
\hline Multiple Index & 1.100 & 1.089 & 1.120 & 1.092 & 1.089 & ${1.060}$   \\
\hline Quadratic Linear & ${1.092}$ & 1.184 & 1.121 & 1.112 & 1.119 & ${1.092}$   \\
\hline Complex Quadratic & 1.232 & 1.337 & 1.283 & 1.301 & 1.277 & ${1.202}$   \\
\hline
\end{tabular}
\caption{Comparison of Deep FDNN and FBNN with early stopping under different settings between the functional predictors and the continuous response.}
\label{at2}
\end{table}

\noindent \footnotesize{Note: (2,2) means two continuous hidden layer with two continuous neurons each and (4,4) means two continuous hidden layer with four continuous neurons each.}

\begin{table}[H]
\centering
\begin{tabular}{|l|l|l|l|l|l|l|l|}
\hline
Mapping & FLM            & CAM            & NN  & FNN        & FDNN           & FBNN                      \\ \hline
Logistic          & {0.544} & {0.544} & 0.591  & 0.562  & {0.548} & {0.546}            \\ \hline
CAM          & {0.663}          & {0.662} & 0.698 & 0.693    & {0.665} & {0.663}                   \\ \hline
Single Index           & 0.665          & 0.666          & 0.680 & 0.673   & {0.663} & {0.662}          \\ \hline
Multiple Index           & 0.661          & {0.660}          & 0.685 & 0.668   & {0.658} & {0.658}    \\ \hline
Quadratic          & 0.639          & 0.595          & 0.652 & 0.615    & {0.584} & {0.596}     \\ \hline
Complex Quadratic          & {0.664}          & {0.663}          & 0.693  & 0.670  & {0.661} & {0.661}  \\ \hline
\end{tabular}
\caption{Comparison of log-likelihood for the methods under different settings between the functional predictors and the binary response.}
\label{tt3}
\end{table}

\section*{Appendix C. Neural Network and Functional Neural Network under similar parameter setting}

We compare the result of different deep learning methods to our approach under the same parameter setting, otherwise for NN and FNN we did a grid search for the number of layers and neurons in the range (1,10). In the same parameter settings, we consider multiple combinations of (continuous) hidden layers $(l=1,2)$ and (continuous) neurons $(j=1,2,4)$. We can see from Table \ref{ann1} and Table \ref{ann2} that NN and FNN preforms worse when we consider a shallow network and our approaches are Superior in both the continuous response and binary response.

\begin{table}[h]
\centering
\begin{tabular}{|l|l|l|l|l|}
\hline
Mapping           & NN    & FNN            & FDNN           & FBNN           \\ \hline
Linear            & 1.201 & 1.193          & 1.019          & {1.011} \\ \hline
CAM               & 1.506 & 1.312          & 1.076          & {1.058} \\ \hline
Single Index      & 1.545 & 1.287          & {1.067} & {1.061} \\ \hline
Multiple Index    & 1.611 & 1.320          & 1.100          & {1.060} \\ \hline
Quadratic Linear  & 1.536 & 1.276          & {1.092} & {1.092} \\ \hline
Complex Quadratic & 1.960 & 1.855          & 1.232          & {1.202} \\ \hline
\end{tabular}
\caption{Comparison of Neural Network and Functional Neural Network (RMSE) with our approach under similar parameter setting with functional predictors and continuous response.}
\label{ann1}
\end{table}

\begin{table}[h]
\centering
\begin{tabular}{|l|l|l|l|l|}
\hline
Mapping           & NN    & FNN    & FDNN           & FBNN           \\ \hline
Linear            & 0.292, 0.608 & 0.290, 0.571 & {0.278, 0.548}  & {0.277, 0.546} \\ \hline
CAM               & 0.407, 0.723 & 0.401, 0.701 & {0.393, 0.665} & {0.392, 0.663} \\ \hline
Single Index      & 0.379, 0.702 & 0.395, 0.689 & {0.357, 0.663} & {0.355, 0.662} \\ \hline
Multiple Index    & 0.391, 0.709 & 0.416, 0.692 & {0.370, 0.658} & {0.368, 0.658} \\ \hline
Quadratic Linear  & 0.344, 0.686 & 0.349, 0.627 & {0.322, 0.584} & {0.324}, 0.596 \\ \hline
Complex Quadratic & 0.408, 0.729 & 0.405, 0.711 & {0.392, 0.661} & {0.393, 0.661} \\ \hline
\end{tabular}
\caption{Comparison of Neural Network and Functional Neural Network (Classification error, Log-likelihood) with our approach under similar parameter setting with functional predictors and Binary response.}
\label{ann2}
\end{table}

\end{document}